\documentclass[lettersize,journal]{IEEEtran}
\usepackage{amsmath,amsfonts}
\usepackage{algorithmic}
\usepackage{array}
\usepackage[caption=false,font=normalsize,labelfont=sf,textfont=sf]{subfig}
\usepackage{textcomp}
\usepackage{stfloats}
\usepackage{url}
\usepackage{verbatim}
\usepackage{graphicx}
\usepackage{cleveref}
\usepackage[export]{adjustbox}
\def\BibTeX{{\rm B\kern-.05em{\sc i\kern-.025em b}\kern-.08em
    T\kern-.1667em\lower.7ex\hbox{E}\kern-.125emX}}
\usepackage{balance}

\def\mybn{\texttt{IN}}
\def\myrelu{\texttt{ReLU}}
\def\myconv{\texttt{Conv}}
\def\mypreb{\texttt{PreB}}
\def\mypost{\texttt{PostB}}
\def\mymaxpool{\texttt{MaxPool}}
\def\myTconv{\texttt{TConv}}
\def\myreshape{\texttt{Reshape}}
\def\myupsample{\texttt{UpSample}}
\def\capsuleEncoder{\texttt{CE}}
\def\capsuleDecoder{\texttt{CD}}
\def\doubleblockdown{\texttt{DBD}}
\def\doubleblockup{\texttt{DBU}}
\def\ColQuantBlock{\texttt{CQB}}
\def\TempColQuantBlock{\texttt{TCQB}}
\begin{document}
\title{UW-ProCCaps: UnderWater Progressive Colourisation with Capsules}
\author{ Rita Pucci, Niki Martinel
\thanks{Manuscript created October 2022; }}

\markboth{Journal of \LaTeX\ Class Files,~Vol.~18, No.~9, September~2020}%
{How to Use the IEEEtran \LaTeX \ Templates}

\maketitle

\begin{abstract}
Underwater images  are fundamental for studying and understanding the status of marine life. We focus on reducing the memory space required for image storage while the memory space consumption in the collecting phase limits the time lasting of this phase leading to the need for more image collection campaigns. We present a novel machine-learning model that reconstructs the colours of underwater images from their luminescence channel, thus saving 2/3 of the available storage space. Our model specialises in underwater colour reconstruction and consists of an encoder-decoder architecture. The encoder is composed of a convolutional encoder and a parallel specialised classifier trained with webly-supervised data. The encoder and the decoder use layers of capsules to capture the features of the entities in the image. The colour reconstruction process recalls the progressive and the generative adversarial training procedures. The progressive training gives the ground for a generative adversarial routine focused on the refining of colours giving the image bright and saturated colours which bring the image back to life. We validate the model both qualitatively and quantitatively on four benchmark datasets. This is the first attempt at colour reconstruction in greyscale underwater images.
Extensive results on four benchmark datasets demonstrate that our solution outperforms state-of-the-art (SOTA) solutions ..... We also demonstrate that the generated colourisation enhances the quality of images compared to enhancement models at the SOTA.
\end{abstract}


\section{Introduction}\label{sec1}
\IEEEPARstart{T}{he} oceans cover most of our planet and are as fascinating as harsh, complex, and dangerous for exploration. Part of the exploration and protection of the rich ecosystems leverages images and videos collected by sophisticated visual sensing systems that allow biologists to analyse them in safe lab environments. These systems are often embedded in robots -- an attractive option because of their non-intrusive, passive, and energy-efficient nature.
Such a solution has been used for monitoring the coral barrier reef \cite{shkurti2012multi}, exploring the depth of the ocean \cite{whitcomb2000advances}, analysing the seabed \cite{bingham2010robotic}, and much more.
However, the application of robots comes with limitations in terms of power and memory storage capacity.
In this work, we focus on reducing the memory space needed to store the collected data. 
This is achieved by storing the greyscale version of the acquired image and restoring the colours when the collection phase is over. 
This is achieved by proposing a novel architecture for automatic image colourisation. Before diving into the innovations proposed in this paper we introduce the challenge posed by the colourisation problem. 

The image colourisation task is a challenging and ill-posed problem because the "correct" colourisation of an entity is not always well defined. There are entities that have a small range of colours, e.g., the grass is usually a tone of green, clouds are generally white, and the sky is blue or black, but others have a broader variation in appearance, e.g., t-shirts, desks, and many other objects which can appear in different shapes and colours. 
\begin{figure}[!t]
\begin{center}
\includegraphics[width=\linewidth]{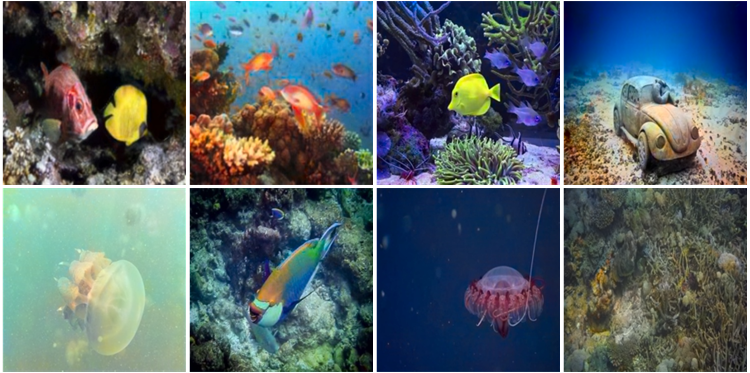}
\caption{Samples of colours generated with UW-ProCCaps for underwater greyscale images.}
\label{fig1}
\end{center}
\end{figure}
Recent works model the global content (image-level features) and objects' instances (entity-level features) in order to extract the needed features to deal with the multimodality of colourisation. Works like~\cite{cheng2015deep,iizuka2016let,zhang2016colorful,larsson2016learning,deshpande2017learning,guadarrama2017pixcolor,isola2017image,he2018deep,DeOldify} focus mainly on the global content resulting in unsaturated entities' colourisation with smudges on boundaries.
Works presented by~\cite{ozbulak2019image,mouzon2019joint,Su2020CVPR} extract both types of features but neglect the importance of the interaction between the global content and the objects' instances for the colourisation of an image. \textit{Differently}, in \cite{pucci2022pro}, we proposed a model that encourages the collaboration between the entities and the global content of the image increasing the naturalness of generated colours. 

In this paper, we propose UW-ProCCaps an encoder-decoder architecture. The UW-ProCCaps takes as input a greyscale image reducing the dimension of the store image to a third of the original dimension, the two coloured channels are not memorised. The architecture consists of convolutional and capsule layers to extract the image-level and the entity-level features respectively. The colour channels are reconstructed through the collaboration of the encoder and decoder which are connected by skip connections. The UW-ProCCaps is able to deal with the multimodality of colourisation by focusing the attention on the entity to be colourised and by reconstructing the colours based on the structural information given by the encoder phase. The architecture is inspired on~\cite{pucci2022pro} and it is specialised in underwater colour generation. In Fig.~\ref{fig1}, we show samples of colours generated with our architecture.
We compare UW-ProCCaps, with two works that represent the SOTA of the two types of feature extraction, Deodify~\cite{DeOldify}, InstanceAware~\cite{Su2020CVPR}. The comparison is presented qualitatively and quantitatively on PSNR, SSIM, and LPIPS metrics over four benchmarks. We demonstrate that UW-ProCCaps is able to generate saturated and vibrant colours for underwater images from their luminescence and it outperforms the stat-of-the-art for the metrics.

\section{Related Work}\label{sec2}
\subsection{Underwater image reconstruction}\label{sec2.1}
\begin{figure}[!t]
\begin{center}
\includegraphics[width=\linewidth]{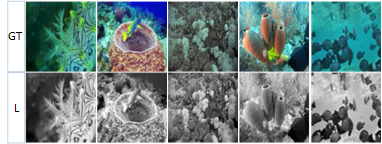}
\caption{Samples of underwater images. The first row consists of coloured images from the benchmarks, the second row consists of the luminescence channel extracted by the images.}
\label{fig2}
\end{center}
\end{figure}

Underwater images are characterised by a whole range of light distortions due to the water absorption of light waves~\cite{akkaynak2017space,smith1978optical,rasmus2004optical}. The colours and the edges' perception change with depth, illumination, and turbidity of water making the image appear completely different. The low definition of edges, the distortion of colours in a blueish or greenish colourisation, and the distortion of light effects mainly the luminescence channel of the images. Fig.~\ref{fig2} shows the original image from the benchmark and the greyscale image obtained with the \textit{L} channel.

\subsection{Colours reconstruction}\label{sec2.2}
The automatic image colourisation task is an open challenge that is generally addressed following two main directions. The former considers image-level features, hence the colourisation is done without specifically considering object entities. In this direction~\cite{isola2017image} with the application of conditional adversarial networks for an image-to-image translation, or~\cite{nazeri2018image, vitoria2020chromagan} with generative models by predicting with semantic and prior knowledge, or ~\cite{zhang2016colorful,larsson2016learning,zhao2018pixel,zhang2017real} with multi-task models and single-pixel significance to predict a right colourisation for the image. 
Methods following the latter direction consider entity-level features. These are introduced by semantic labels as in ~\cite{iizuka2016let,zhao2018pixel}. These features are interpretable semantics by a cross-channel encoding scheme~\cite{zhang2016colorful} and then enriched with a pre-trained classification model~\cite{mouzon2019joint}. \cite{Su2020CVPR} implements the entity-level and image-level features extractions with two parallel models. In \cite{DeOldify}, the focus is on the image-level features with an encoder-decoder structure trained in two phases (end-to-end, then GAN).

Our model integrates the entity-level and image-level feature extractions in one architecture, trained in two phases. The former follows the progressive learning paradigm to let the architecture gradually learn how to reconstruct colours. The latter exploits a GAN training scheme to improve the naturalness of the predicted colours.

\subsection{Progressive learning}\label{sec2.3}
ProGL is a training methodology proposed by~\cite{karras2017progressive} for generative networks. It consists of starting with low-resolution images, and then progressively increasing the resolution by adding layers. This allows the training to first discover the large-scale structure of the image distribution and then shift attention to the fine details. Works like \cite{karras2017progressive, natsumi2019} proposed a ProGL scheme for GANs to generate images of people, \cite{kang2020pl} applied ProGL to obtain a multi-tasking method based on salient object detection. 
A first application of progressive learning (ProGL) for colour reconstruction was proposed in~\cite{pucci2022pro}.
In our work, ProGL is applied in the first phase of training to let the model learn gradually the collaboration among features extracted by the different layers to reconstruct the colour information.
\section{Proposed method}\label{sec3}
We consider the images in the CIELab colour space (where the luminescence channel \textit{L} is independent of the chrominance channels \textit{(a,b)} which identify the four unique colours of human vision) with the assumption that only the  luminescence channel \textit{L} is stored while recording underwater videos.

\subsection{UW-ProCCaps architecture}\label{sec3.1}
\begin{figure*}[!t]
\includegraphics[width=\linewidth]{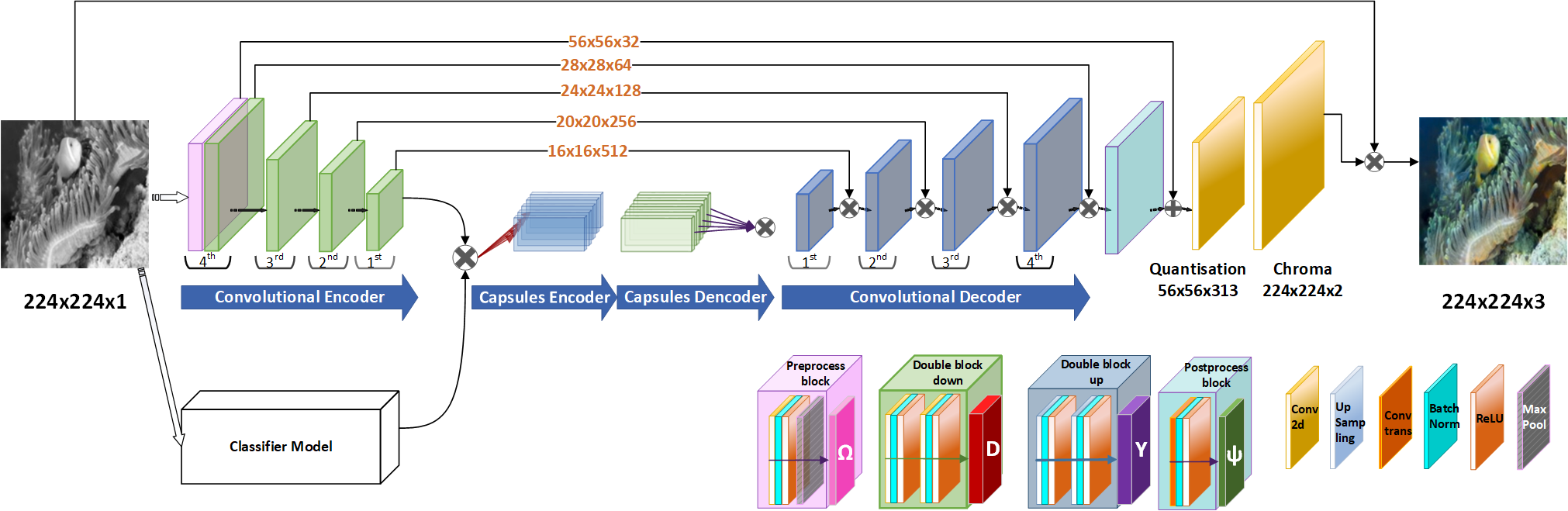}
\caption{Proposed UW-ProCCaps architecture. The classification module is pre-trained and frozen while the entire architecture is trained first end-to-end and then in GAN to reconstruct the chromatic channels of underwater images.}
\label{fig3}
\end{figure*}
The UW-ProCCaps architecture follows an encoder-decoder structure. The encoder is composed of two parallel models, a convolutional encoder, and a classifier. The former processes the inputs by transforming them into a set of feature maps. The latter emits classification vectors. Feature maps and classification vectors contain image-level information. A capsules encoder aggregates such vectors to emit entity-level features. 
The decoder, composed of a capsules decoder and a convolutional decoder, receives the entity-level features to generate the chromatic channels. 
The convolutional encoder and the convolutional decoder implement a tight collaboration by skip and residual connections which connect each layer of the convolutional encoder with the correspondent layer in the convolutional decoder, as illustrated in Fig.~\ref{fig3}.

\paragraph{Convolutional Encoder}
The luminescence channel of the input image, $\mathbf{I}_{L}\in\mathbb{R}^{224\times 224\times 1}$, is given to the convolutional encoder and to the classifier. The convolutional encoder consists of the combination of a preprocessing block (\mypreb) and $n$ double blocks down ($\doubleblockdown^n$ where $n\in[4,..,1]$). The \mypreb{ }is composed of a \myconv $-$\mybn $-$\myrelu $-$\mymaxpool. It performs an initial features extraction yielding $\mathbf{\Omega}=f_{\mypreb}(\mathbf{I}_{L})\in\mathbb{R}^{56\times 56\times 32}$.
The $\mathbf{\Omega}$ goes through all the \doubleblockdown{ }s, each of which is composed of two consecutive sequences of \myconv$-$\mybn$-$\myrelu, \cite{hadji2018we}. \myconv{ }is a $3\times3$ heterogeneous convolution;  $\mybn$ is the instance normalisation; and \myrelu{ } is used to non-linearly transform the fused features. 
At its last stage, $\doubleblockdown^1$ outputs $\mathbf{D}^{1}\in\mathbb{R}^{16\times 16\times 512}$, i.e.,
\begin{equation}
\mathbf{D}^1=f_{\doubleblockdown^{2}}(f_{\doubleblockdown^{3}}(f_{\doubleblockdown^{4}}(f_{\mypreb}(\mathbf{I}_{L})))
\end{equation}

\paragraph{Classifier Model}
The classifier model enriches the information extracted by the convolutional encoder with specific features related to the predicted class of the image and helps the model deal with the multimodality of the colorization by clustering the information by class. 
The classifier provides a classification vector formatted to be concatenated with $\mathbf{D}^{1}$ in $\mathbf{\Upsilon}\in\mathbb{R}^{16\times 16\times 519}$, explained in Sec.\ref{sec3.2}. 

\paragraph{Capsules Encoder (\capsuleEncoder)}
This encoder aggregates the image-level information included in $\mathbf{\Upsilon}$ to extract entity-level features. 
This is achieved in three steps. 
We first compute the activation vectors $\mathbf{U}$: 
\begin{equation}
\mathbf{U} = [Flatten(\myconv^1(\mathbf{\Upsilon}))^T, ..., Flatten(\myconv^C(\mathbf{\Upsilon}))^T] 
\end{equation}
where $C=32$ is the number of capsules and each column of $\mathbf{U} \in \mathbb{R}^{k\times C}$ is the capsule output $u_{i}\in \mathbb{R}^k$. The second step is to compute the entity-level prediction vectors $\hat{\mathbf{U}}$. We apply an affine transformation on $\mathbf{U}$ with a weight matrix $W_{ij}\in\mathbb{R}^{k\times k}$ to obtain
\begin{equation}
   \hat{\mathbf{u}}_{j|i}=\textbf{W}_{ij}\mathbf{u}_i
\end{equation}
In the third step, we apply the ``routing by agreement'' mechanism \cite{sabour2017dynamic} on $\hat{\mathbf{U}}$. This mechanism, during the training phase, uses the coupling coefficients $\mathbf{c}_{i|j}$ to identify the clusters of the features in $\hat{\mathbf{U}}$. Each cluster of features identifies one (or part of an) entity and is formally defined by the weighted sum of $\hat{\mathbf{u}}_{j|i}$ vectors:
\begin{equation}
    \mathbf{v}_j = squash(\sum{\mathbf{c}_{i|j}*\hat{\mathbf{u}}_{j|i}})
\end{equation}
where $\mathbf{v}_j \in \mathbb{R}^{\hat{k}}$. The final output of downsampling is the matrix $\mathbf{V}={\mathbf{v}_0,\cdots, \mathbf{v}_j}$, it carries information about how strong the capsules agree on the presence of an entity. $\mathbf{V}\in\mathbb{R}^{32\times 8\times 8\times 128}$ feature matrix for each capsule contains the entity-level features matrix.

\paragraph{Capsules Decoder}
The capsules decoder (\capsuleDecoder) elaborates the entity-level features $\mathbf{V}$ to reconstruct the colours' information.
The features in $\mathbf{V}$ lack information about their spatial displacement with respect to the input datum. This is however needed to properly reconstruct colours within entities' boundaries.
The \capsuleDecoder{ }inverts the \capsuleEncoder{}{ } process.
A weight matrix $\mathbf{W}^{r}_{ji} \in \mathbb{R}^{\hat{k}\times k}$ reverses the affine transformation:
\begin{equation}
\mathbf{u}^{r}_{i|j} = \mathbf{W}^r_{ji}\mathbf{v}_j    
\end{equation}
then $\mathbf{u}{r}_{i|j}\in \mathbb{R}^k$ are stacked in $\mathbf{U}^r \in \mathbb{R}^{k\times C}$. $\mathbf{u}^{r}_{i}$'s are given to the $i$-th reversed capsules, implemented as a transpose convolutional layer ($\myTconv_i$).
This yields to
\begin{multline}
    \mathbf{X} = [\myreshape(\myTconv_1(\mathbf{{u}}^{r}_1)),\cdots,\\
    \myreshape(\myTconv_k(\mathbf{{u}}^{r}_k))]
\end{multline}
where $\mathbf{{u}}^{r}_i$ denotes the $i$-th row of $\mathbf{U}^r$. 
The matrix $\mathbf{X}$ consists of the initial colours reconstruction from entity-level features $\mathbf{V}$. 

\paragraph{Convolutional Decoder}
$\mathbf{X}$ is the input to the convolutional decoder which consists of four $\doubleblockup$ stacked layers.
DIRE COME E' FATTO DBU
At its first stage, i.e., $\doubleblockup^{1}$, the block input is $\mathbf{X}$, which generates:
\begin{equation}
  \mathbf{Y}^{1} =f_{\doubleblockup^{1}}(\mathbf{X})
\end{equation}
Following blocks ($\doubleblockup^{m}$, with $m \in [2,..,4]$), apply a skip connection to promote the collaboration between the encoder and decoder phases, i.e., 
\begin{equation}
  \mathbf{Y}^{m} =f_{\doubleblockup^{m}}(cat(\mathbf{Y}^{m-1},\mathbf{D}^{m-1}))
\end{equation}
The stacked $\doubleblockup$ layers are followed by the last upsampling layer (\mypost).
This layer performs the reversed function of \mypreb{ }:
\begin{equation}
  \mathbf{\Psi} = f_{\mypost}(cat(\mathbf{Y}^{4},\mathbf{D}^{4}))  
\end{equation}
The \mypost{ } outputs $\mathbf{\Psi}\in\mathbf{R}^{H\times W\times \Gamma}$ having the same spatial resolution of $\mathbf{\Omega}$. 
It consists of the final composition of all the features extracted in the \myupsample{ }phase. The five functional blocks consist of the 28-layer network.

\paragraph{Quantization of colours}
The last two layers of UW-ProCCaps are two convolutional layers which quantize and reconstruct the colours predicted in $\mathbf{\Psi}$, we refer to this block as the colour quantisation block (\ColQuantBlock). The first convolutional layer computes a quantised representation of the colours, based on the idea in~\cite{zhang2016colorful}. This prevents the model to generate values outside the set of gamut colours, which would provide implausible results as demonstrated in \cite{pucci2022pro}. Following CIT, $\mathbf{\Psi}$ is remapped and quantised in the in-gamut CIELab colours with $bin=10$ to obtain 313 colour classes. 
This layer receives the residual of $\mathbf{\Psi}$ and $\mathbf{\Omega}$ and maps it over the quantised colour distribution
\begin{equation}
    \mathbf{\hat{Z}} = f_{Quantisation}(sum(\mathbf{\Psi},\mathbf{\Omega}))
\end{equation}
where $\mathbf{\hat{Z}}\in\mathbb{R}^{56\times 56\times 313}$. This makes the task a classification problem for each point in the input. 
The second convolutional layer computes the chroma representation which lets the model predict the $a$ and $b$ channels consistent with the CIELab colour definition. This layer consists of a $1\times 1$-\myconv{ } layer followed by bilinear upsampling to resize $\mathbf{\hat{Z}}$ by a factor of 4, hence to map $\mathbf{\hat{Z}}$ onto the two chrominance channels $(\hat{a},\hat{b})\in\mathbb{R}^{224\times 224\times 2}$.

\subsection{Loss function}\label{sec3.2}
The UW-ProCCaps model training consists of three phases, described in Sec.~\ref{sec3.3}. In each phase, we apply the loss function relevant to the corresponding task and training methodology. \textbf{Classifier finetuning}: The classifier receives the  $\mathbf{I}_{L}\times 3$ in input and provides the classification of the input image $\hat{y}\in {c_0,..., c_p}$ where $p$ is the number of classes in the training dataset. In this phase, the class of the input image is known as $y$. The loss function is computed as $\mathcal{L}_{class} = CrossEntropyLoss(\hat{y},y)$. \textbf{End-to-End UW-ProCCaps training}: The UW-ProCCaps is trained progressively and we compute a composed loss function which takes into consideration the two layers of reconstruction that is in the \ColQuantBlock, $\mathcal{L}_{end2end} = \mathcal{L}_q+\mathcal{L}_ch$. Where the $\mathcal{L}_q$ is the quantised colours loss and $\mathcal{L}_ch$ is the chrominance loss. To compute the $\mathcal{L}_q$, the matrix $\mathbf{\hat{Z}}$ is compared with the projection of ground-truth chroma channels on the quantised representation \cite{zhang2016colorful}. The ground truth, $\mathbf{I}_{a,b}$, is converted by the soft-encoding scheme in the quantised representation $\mathbf{Z}$.
\begin{equation}
    \mathcal{L}_{q} = -\sum_{h,w}v(\mathbf{Z}_{h,w}) \sum_{q}\mathbf{Z}_{h,w,q}log(\mathbf{\hat{Z}}_{h,w,q})
    \label{eq:lq}
\end{equation}
where $v(\cdot)$ re-weights the loss for each pixel based on pixel colour rarity. We have considered the soft-encoding and the $v(\cdot)$ values introduced by~\cite{zhang2016colorful}.
To compute the $\mathcal{L}_ch$, we minimise the difference between the real ($\mathbf{I}_{a,b}$) and the predicted ($\mathbf{I}_{\hat{a},\hat{b}}$) colour channels as:
\begin{equation}
\mathcal{L}_{ch} = ||\hat{a}-a||^2_2 + ||\hat{b}-b||^2_2.
\label{eq:lc}
\end{equation}
\textbf{GAN UW-ProCCaps Training}:
In the GAN phase of training, we apply a composed loss function $\mathcal{L}_{GAN} = \mathcal{L}_{ADV}+\mathcal{L}_{perc}$ where $\mathcal{L}_{ADV}$ is the adversarial loss ~\cite{goodfellow2020generative} and $\mathcal{L}_{perc}$ is the perceptual loss~\cite{suvorov2021resolution}. The adversarial loss is implemented with the Binary Crossentropy (BCE) with logits. The perceptual loss evaluates the distance between features extracted from the predicted colourisation and the ground truth images by a base pre-trained network $\phi(\cdot)$. It does not require an exact reconstruction, allowing for variations in the
reconstructed image and focused on an understanding of global structure. With this purpose the $\mathcal{L}_{perc}$ uses a high receptive field base model:
\begin{equation}
\mathcal{L}_{perc} = M([\phi_{HRF}(\mathbf{Z})-\phi_{HRF}(\mathbf{\hat{Z}})]^2)
\end{equation}
$M$ is the sequential two-stage mean operation (interlayer mean of intralayer means). The $\phi_{HRF}(x)$ is implemented using Dilated convolutions~\cite{suvorov2021resolution}. 

\begin{figure*}[!t]
\begin{center}
\includegraphics[width=\linewidth]{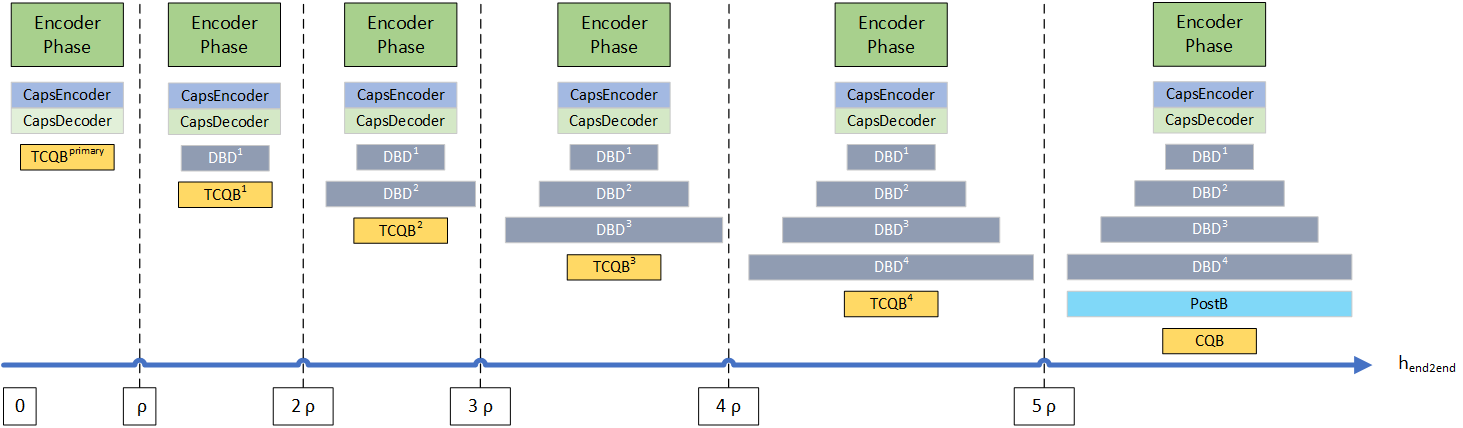}
\caption{The ProGL method is applied in this paper over the reconstruction phase to give the model the time to consolidate the knowledge extracted at each layer of reconstruction.}
\label{fig4}  
\end{center}
\end{figure*}

\subsection{Training UW-ProCCaps}\label{sec3.3}
We train the model in three phases. 
\textbf{Classifier finetuning}: The classifier used for this work is available at the stat-of-the-art. We use the pre-trained model and we fine-tune it on the Flickr-UW-7(described in Sec.~\ref{sec4.1} dataset for $h_{class}$ epochs. This initial phase lets the classifier create an underwater image structure understanding. 

\begin{table*}[t]
\begin{center}
\caption{Focus on the output images obtained at each step of progression performed while training in the \textbf{End-to-End UW-ProCCaps training phase}~\ref{sec3.2} }
\label{tab:3.2.1}
\begin{tabular}{|l|c|c|c|c|c|c|c|}
\hline
&\capsuleDecoder&$\doubleblockup^1$&$\doubleblockup^2$&$\doubleblockup^3$&$\doubleblockup^4$&Output\\
 $\mathbf{Y}^m$&$15\times 15$&$16\times 16$&$20\times 20$&$24\times 24$&$28\times 28$&$224\times 224$\\
\hline
\end{tabular}
\end{center}
\end{table*}
\textbf{End-to-End UW-ProCCaps training}: The entire UW-ProCCaps architecture described in Sec.~\ref{sec3.1}, except for the classifier model which is frozen, is trained on UFO120 dataset for $h_{end2end}$ epochs following the ProGL methodology presented in Sec.~\ref{sec2.3}. To implement the progression we introduce the temporary quantisation block (\TempColQuantBlock) which consists of a  quantisation layer and a chroma layer in which dimensions of the output change in based on the depth of the step, as explained in Tab.\ref{tab:3.2.1}. The progression methodology trains the model adding a \doubleblockup{ }every $\rho$ epochs of training. Together with the $\doubleblockup^{m}$, ProGL adds the relative $\TempColQuantBlock^{m}$, and, if it is present, removes the $\TempColQuantBlock^{m-1}$, as shown in Fig.~\ref{fig4}. 
Each $\TempColQuantBlock^{m}$ follows the structure proposed for \ColQuantBlock, where the resolution of $\hat{\mathbf{Z}}$ and $(\hat{a},\hat{b})$ is equal to $\mathbf{Y}^m$, and it is defined by the level of progression, Tab.~\ref{tab:3.2.1}. At the beginning of training, the \myupsample{ }consists only of \capsuleDecoder. Let \capsuleDecoder{ }being the first layer of reconstruction, $\TempColQuantBlock^p$ provides $\mathbf{\hat{Z}}^{p}$ and $(\hat{a},\hat{b})^{p}$. In following layers, for $\doubleblockup^{m}$, we add $\TempColQuantBlock^{m}$ and it provides $\mathbf{\hat{Z}}^{m}$ and $(\hat{a},\hat{b})^{m}$. The last layer of growing is the \mypost{ }that completes the structure.  

\textbf{GAN UW-ProCCaps Training}: We refine the knowledge of the model by fine-tuning the architecture (while keeping the classifier block frozen) through a GAN training procedure with the Pix2Pix Discriminator~\cite{isola2017image}. The model is trained for $h_{GAN}$ epochs on the UFO120 dataset. We observe that by performing the GAN training phase, the model learns to provide colours more vibrant and neat moreover it results in a greater range of colourisation compare to the model obtained by the End-to-End UW-ProCCaps training phase. This point is discussed in the ablation study at Sec.~\ref{sec4.4}.

\section{Experimental Results}\label{sec4}
\subsection{Datasets}\label{sec4.1}
\paragraph{Training phase} The encoder phase has a parallel classifier. We train such a classifier following a webly-supervised procedure. To do this, we have collected a new dataset, named Flickr-UW-7, by scraping the Flickr platform for free public license images containing one of 7 tags:  dive, coral, fish, jellyfish, seabed, shark-whales, and weirdo. 
The scraping process generated 2009 images. 

The End-to-End UW-ProCCaps and the GAN UW-ProCCaps training phases exploits data in the UFO120 dataset~\cite{islam2020simultaneous}. This contains 1500 samples shot underwater with no labels. Each sample consists of a noisy and denoised image pair. The noisy image is shot underwater and it shows water distortion, while the denoised image does not have water distortion. We considered the noisy images to let the model learn to colourise the original underwater images. 

\paragraph{Validation phase} We evaluated our model on four benchmarks. 
In the \textbf{Enhancing Underwater Visual Perception (EUVP)} dataset~\cite{islam2020fast}, we considered the validation split consisting of 515 paired images with water distortion.
The Heron Island Coral Reef Dataset (HICRD)~\cite{CSIRO} is focused on the coral reef in the deep sea, we use the paired HR split which consists of 300 images.
The Underwater Image Enhancement Benchmark (UIEB)~\cite{peng2021ushape} includes 890 images, which involve rich underwater scenes (lighting conditions, water types, and target categories) and better visual quality reference images than the existing ones.
The Underwater Image Super-Resolution (USR248)~\cite{islam2020simultaneous} contains 248 samples of underwater images.

\subsection{Implementation details}\label{sec4.2}
The input images $\mathbf{I}_{L}\in \mathbb{R}^{224\times 224\times 1}$ and the outputs of UW-ProCCaps are $\mathbf{\hat{Z}}\in \mathbb{R}^{56\times 56\times 313}$ and the $(\hat{a},\hat{b})\in \mathbb{R}^{224\times 224\times 2}$. We implemented the classification model with a ResNet34~\cite{he2016deep} (hereafter referred to as ResNet). In the training process, we used a batch of 16 samples and the Adam optimiser with a learning rate of $2e^{-3}$. In the classifier training, we trained the ResNet for $h_{class} = 20$ epochs. In the End-to-End training, we set $\rho=30$, hence run $h_{end2end} = 240$ epochs in ProGL. Then we trained the whole network in GAN for $h_{GAN} = 1000$ epochs.

\subsection{Evaluations metrics}\label{sec4.3}
The UW-ProCCaps model is evaluated qualitatively for the naturalness of the predicted colourisation, and quantitatively over the metrics results. To assess our colourisation performance, we follow the experimental protocol in~\cite{Lei_2019_CVPR} and consider the Peak Signal to Noise Ratio (PSNR), the Learned Perceptual Image Patch Similarity (LPIPS)~\cite{zhang2018perceptual} (version 0.1 with VGG backbone), and the Structural Similarity Index Measure (SSIM)~\cite{wang2004image}.

\subsection{Ablation study}\label{sec4.4}
The proposed architecture consists of different parts and implementation choices joined together to perform the task of colourisation. In this section, we analyse the importance of each of the main parts proposed.
\paragraph{Are the capsule bringing a beneficial outcome?} To answer this question we present quantitative and qualitative results obtained with and without capsules over the validation datasets introduced in Sec.~\ref{sec4.1}. The UW-Net consists of the encoder, the convolutional decoder, and the quantisation of colours as described in Sec.~\ref{sec3.1}. Both networks are trained following the ProGL methodology for only the End-to-End training phase.
\begin{figure}[!t]
\begin{center}
\includegraphics[scale=.50]{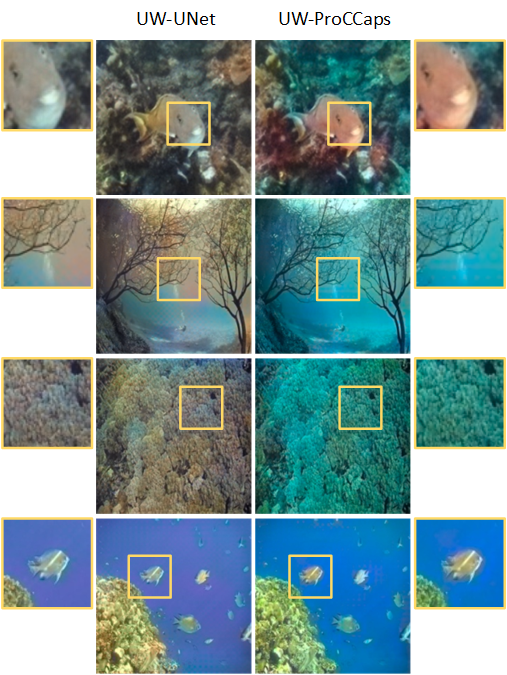}
\caption{Ablation study a): The important role of the capsules in the UW-ProCCaps is here proved by taking a closer look at some details of the colour reconstruction obtained with and without Capsules.}
\label{fig5}  
\end{center}
\end{figure}
Fig.~\ref{fig5} shows the results of colour reconstructions obtained with UW-UNet and UW-ProCCaps. 
UW-ProCCaps reconstructs pleasant colourisation that looks natural, plausible, and well defined in the contours of the entities in the images. In Tab.~\ref{tab1}, we compare the metrics results obtained with the UW-UNet and the UW-ProCCaps (referred to in Tab.~\ref{tab1} with the name UWPCC\_FTFlickrResNet\_E2E to be distinguished by the other ablation study cases). We observe that the quantitative results underline that the two networks are competitive with each other. We consider both the qualitatively and quantitatively results to prove that the application of capsules is an important addition to the architecture to obtain a good colour reconstruction.
\paragraph{Is the parallel classifier bringing a beneficial outcome?} To prove the importance of the classifier network, Fig.\ref{fig6} we compare the first row obtained with UW-ProCCaps and the second row obtained with the same architecture but without the classifier network (here referred to as ResNet as described in Sec.~\ref{sec4.2} and both variations are trained following Sec.~\ref{sec3.1}. The colours obtained with UW-ProCCaps are bright and vibrant and they vary from subject to subject, while removing the classifier network the colours are low tone and sometimes are greyish. In Tab.~\ref{tab2} the results on metrics depend on the dataset considered but are always competitive with both models. We propose the UW-ProCCaps with the classifier network considering both the qualitative results and the quantitative results.
\begin{figure}
\begin{center}
\includegraphics[scale=.40]{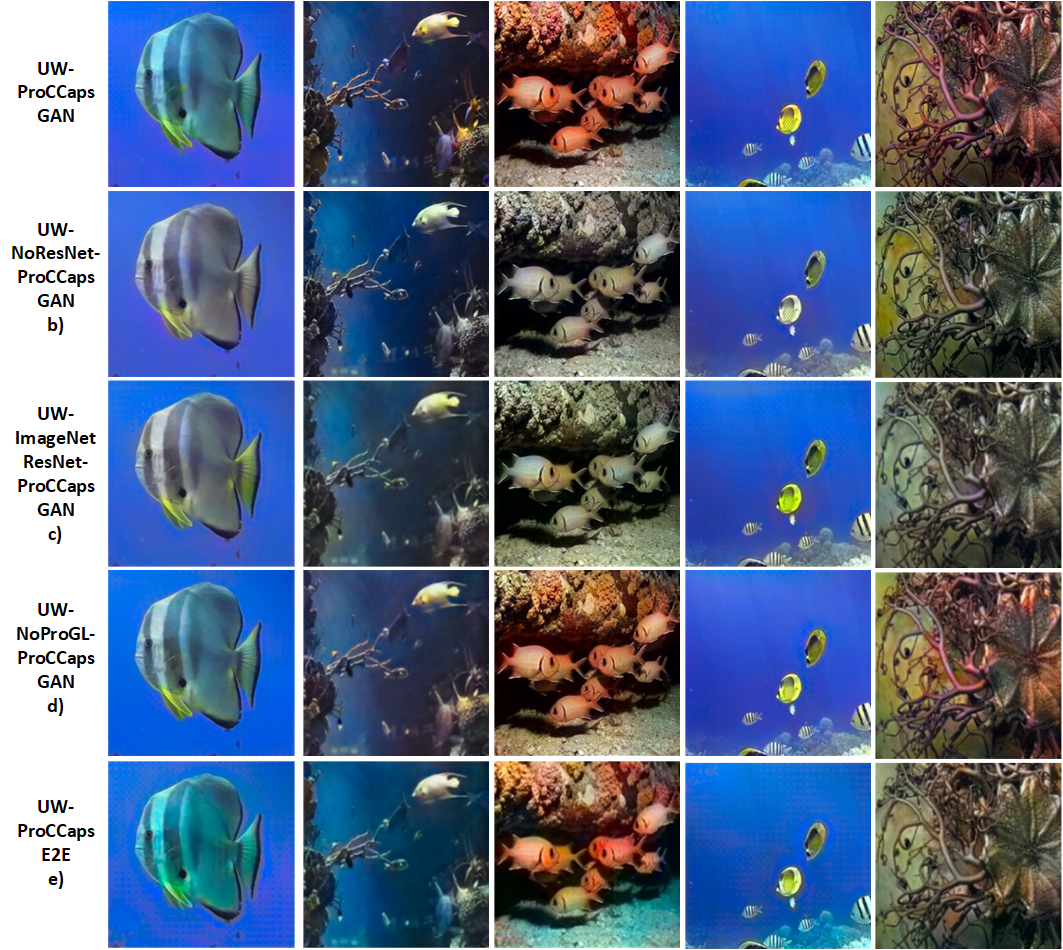}
\caption{Ablation study b,c,d,e): We summarise in this figure the images achieved with the model studied at the ablation study. Each row refers to the points at the Sec.~\ref{sec4.4}}
\label{fig6}
\end{center}
\end{figure}
\paragraph{Do we have to fine-tune the classifier network even if it is already pre-trained?} This question arises naturally and the idea behind the fine-tuning is to let the network focus on the classes that are present underwater improving the model for better colour reconstruction. We perform experiments with the ResNet pre-trained on ImageNet and with the same network fine-tuned on Flickr-UW-7. In Fig.~\ref{fig6}, the third row shows the results obtained with UW-ProCCaps with the ResNet only pre-trained on ImageNet while the first row shows results obtained fine-tuning the ResNet. In the first row, we note that all the entities (fish, corals, and alga) in the images are well-coloured with plausible and diversified colours in contrast with the third row where the majority of entities are not properly coloured. In Tabs.~\cref{tab1,tab2} respectively for end-to-end and GAN, the metrics prove that the fine-tuning of the ResNet improves the quantitative performances with all the datasets considered and for almost all the metrics. 
\paragraph{The progressive learning is applied in the end-to-end training phase, but is it improving the generated colours?} The fourth row in Fig.~\ref{fig6} shows the results obtained with UW-ProCCaps where the end-to-end training phase is performed without the ProGL methodology. We compare the fourth row with the first row, the output a UW-ProCCaps where the ProGL is applied. The first aspect that we note is that the details of the entities in the images are well-defined and the colourisation proposed for them is bright and of high quality compare with the one obtained without the ProGL in end-to-end phase. We prove that the application of ProGL improves the quality of the output obtained by the network. In Tabs.~\cref{tab1,tab2} respectively for end-to-end and GAN, the results obtained for the metrics are improved in almost all the datasets by applying the finetuning of the model in GAN.

\paragraph{Is the GAN training phase improving the performance?} As described in Sec.~\ref{sec3.2}, we train the entire UW-ProCCaps first end-to-end and then in GAN. In the end-to-end phase, the model is trained to achieve the colourisation of the ground truth and ends up with an averaging the possible colours of each pixel in order to reduce the loss error. This ends with under-toned and brownish colourisation. The proof of this observation is shown in Fig.~\ref{fig6} comparing the first row where UW-ProCCaps is trained in GAN and the last row where UW-ProCCaps is trained end-to-end. The colours reconstructed at the end-to-end phase (referred to in the image as E2E) are brownish and the colour tends to be not vibrant compares to the colours reconstructed after the GAN phase. The quantitative results are presented in Tabs.~\cref{tab1,tab2} respectively for end-to-end and GAN. Results describe that the application of a GAN training phase provides a consistent improvement in the colour reconstruction for almost all the variations of the model in all the validation datasets.

\section{Results}\label{sec5}
In this section, we analyse the qualitatively and quantitatively results obtained with UW-ProCCaps. We compare our model with the two well-known models at the stat-of-the-art, Deodify~\cite{DeOldify} and InstanceAware~\cite{Su2020CVPR} for the task of colourisation. In \cite{DeOldify}, authors presented a two phases training as the one proposed in this paper in Sec.~\ref{sec3.2}, and in \cite{Su2020CVPR}, the model architecture takes into consideration the identification of entity-levels and image-level features that here we implement with capsules. Our model and the models at stat-of-the-art are trained following the training proposed in Sec.~\ref{sec3.2}, and on the same datasets described in Sec.~\ref{sec4.1}.
We finally analyse if the model is able to improve the quality of the colours while reconstructing it. with this intent, we compare the results with models at the SOTA trained on the enhancement task on UFO-120, Deep Sesr~\cite{islam2020simultaneous}, Funie gan~\cite{islam2020fast}, Ugan and Ugan-p~\cite{fabbri2018enhancing}. These methods take the coloured noise image as the input image.
\paragraph{Quantitative comparison}\label{sec5.1}
\begin{table*}[]
\begin{center}
\caption{The table summarises the results obtained after the end-to-end training. The models considered are the ones at the ablation studies compared with the Deoldify, and Instance Aware~\cite{DeOldify,Su2020CVPR}.} 
\label{tab1}
\begin{tabular}{|l|rrr|rrr|rrr|rrr|}
\hline
model & \multicolumn{3}{c}{EUVP\_test} & \multicolumn{3}{c}{USR248} & \multicolumn{3}{c}{UIEB} & \multicolumn{3}{c}{HICRD\_pairedHR} \\
{} &      ssim &    psnr &  lpips &   ssim &    psnr &  lpips &   ssim &    psnr &  lpips &           ssim &    psnr &  lpips \\
\hline
Deoldify    &     0.768 &  28.654 &  0.368 &  0.662 &  28.667 &  0.307 &  0.692 &  28.673 &  0.269 &          0.597 &  28.790 &  0.351 \\
InstanceAware   &     0.744 &  28.362 &  0.401 &  0.723 &  28.822 &  0.288 &  0.748 &  28.653 &  0.284 &          0.743 &  29.585 &  0.298 \\
\hline
UW-ProCCaps    &     0.853 &  \textbf{29.509} &  \textbf{0.166} &  0.856 &  29.585 &  0.180 &  0.842 &  \textbf{29.316} &  0.191 &          0.932 &  30.063 &  0.164 \\
UW-NET       &     0.840 &  28.993 &  0.203 &  \textbf{0.880} &  \textbf{30.023 }&  \textbf{0.174} &  \textbf{0.852} &  29.170 &  0.194 &          0.951 &  30.528 &  0.136 \\
UW-NoResNet-ProCCaps &     \textbf{0.862 }&  29.459 &  0.171 &  0.864 &  29.539 &  0.181 &  0.850 &  29.261 &  \textbf{0.188 }&          \textbf{0.954} &  \textbf{30.611} &  \textbf{0.130} \\
UW-ImageNetResNet-ProCCaps &     0.857 &  29.122 &  0.193 &  0.859 &  29.404 &  0.192 &  0.844 &  29.209 &  0.203 &          0.950 &  30.230 &  0.162 \\
UW-NoProGL-ProCCaps &     0.851 &  29.221 &  0.177 &  0.872 &  29.708 &  0.18 &  0.845 &  29.226 &  0.198 &          0.957 &  30.775 &  0.14 \\
\hline
\end{tabular}
\end{center}
\end{table*}
\begin{table*}[t]
\begin{center}
\caption{The table summarises the results obtained after the fine-tuning with GAN. The models considered are the ones at the ablation studies compared with the Deoldify, and Instance Aware~\cite{DeOldify,Su2020CVPR}.}
\label{tab2}
\begin{tabular}{|l|rrr|rrr|rrr|rrr|}
\hline
model & \multicolumn{3}{c}{EUVP\_test} & \multicolumn{3}{c}{USR248} & \multicolumn{3}{c}{UIEB} & \multicolumn{3}{c}{HICRD\_pairedHR} \\
{} &      ssim &       psnr &     lpips &      ssim &       psnr &     lpips &      ssim &       psnr &     lpips &           ssim &       psnr &     lpips \\
\hline
Deoldify    &  0.757 &  28.566 &  0.435 &  0.684 &  28.802 &  0.331 &  0.700 &  28.708 &  0.319 &       0.618 &  29.508 &  0.382 \\
InstanceAware  &  0.808 &  \textbf{29.677 }&  0.225 &  0.717 &  28.729 &  0.262 &  0.741 &  29.067 &  0.215 &       0.741 &  29.322 &  0.224 \\
\hline
UW-ProCCaps    &  0.846 &  28.748 &  0.192 &  \textbf{0.901} & \textbf{30.498} &  \textbf{0.124} &  0.864 &  28.991 &  0.194 &       0.950 &  30.540 &  0.161 \\
UW-NoResNet-ProCCaps &  \textbf{0.858} &  29.018 &  0.189 &  0.881 &  29.586 &  0.181 &  \textbf{0.876} &  29.254 &  \textbf{0.177} &       \textbf{0.961 }&  \textbf{31.294} &  \textbf{0.154} \\
UW-ImageNetResNet-ProCCaps &     0.851 &  29.038 &  0.184 &  0.881 &  29.619 &  0.177 &  0.873 &  29.254 &  0.173 &           0.96 &  31.02 &  0.154 \\
UW-NoProGL-ProCCaps &     0.854 &  29.136 &  \textbf{0.177} &  0.895 &  30.148 &  0.153 &  0.855 &  \textbf{29.271} &  0.177 &          0.956 &  30.839 &  0.165 \\
\hline

\end{tabular}
\end{center}
\end{table*}
\begin{table}
\caption{The table compares the metrics results obtained UW-ProCCaps and the enhancement methods at the SOTA with EUVP dataset.}
\label{tab3}
\begin{center}
\begin{tabular}{|l|c|c|c|}
\hline
{} & \multicolumn{3}{l}{EUVP\_test} \\
{} &      ssim &    psnr &  lpips \\
\hline
UW-ProCCaps    &  0.846 &  28.748 &  0.192\\
DeepSest &     0.768 &  28.195 &  0.283 \\
Funie    &     0.745 &  28.362 &  0.307 \\
Ugan     &     0.794 &  28.394 &  0.258 \\
Ugan\_p   &     0.798 &  28.549 &  0.257 \\
\hline
\end{tabular}
\end{center}
\end{table}
\begin{figure}[!t]
\includegraphics[width=\linewidth, left]{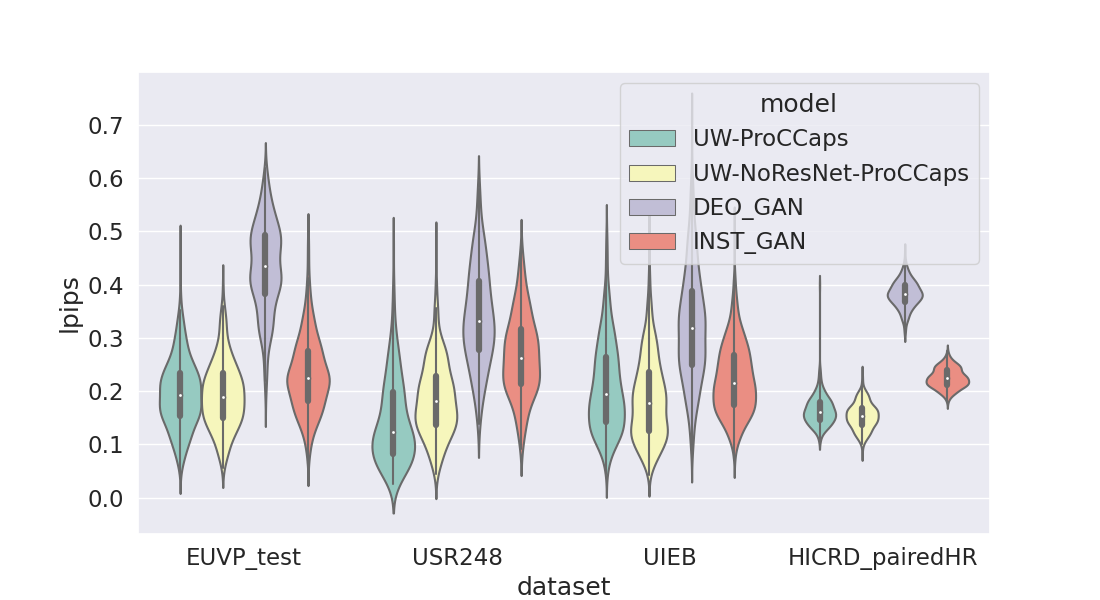}
\caption{Violin plot: compared results for the LPIPS metric.}
\label{LPIPS}
\end{figure}
\begin{figure}[!t]
\begin{center}
\includegraphics[width=\linewidth, left]{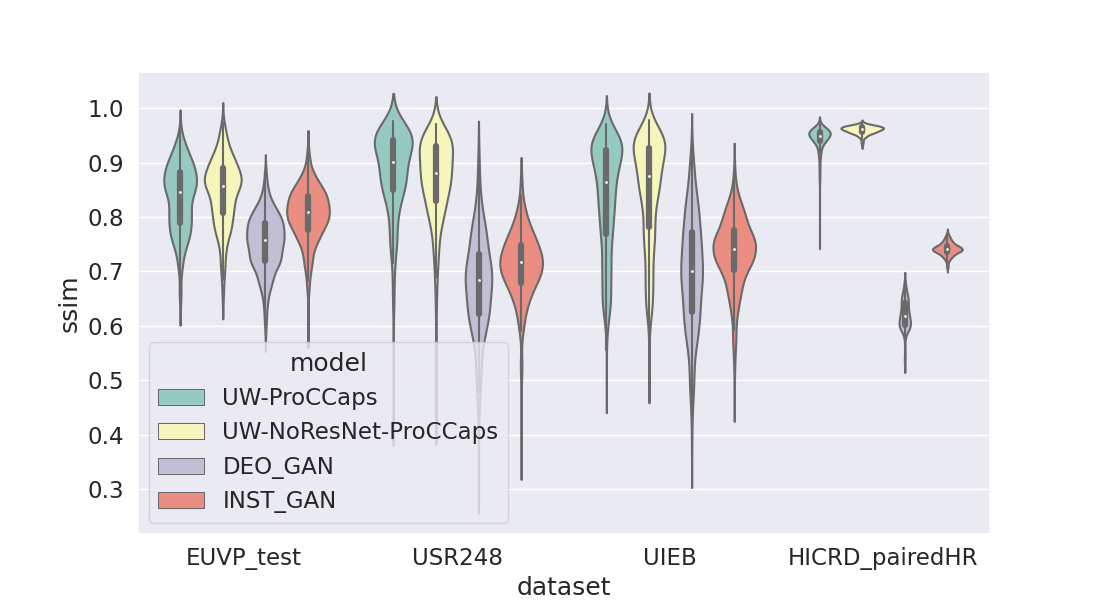}
\caption{Violin plot: compared results for the SSIM metric.}
\label{SSIM}
\end{center}
\end{figure}
\begin{figure}[!t]
\includegraphics[width=\linewidth, left]{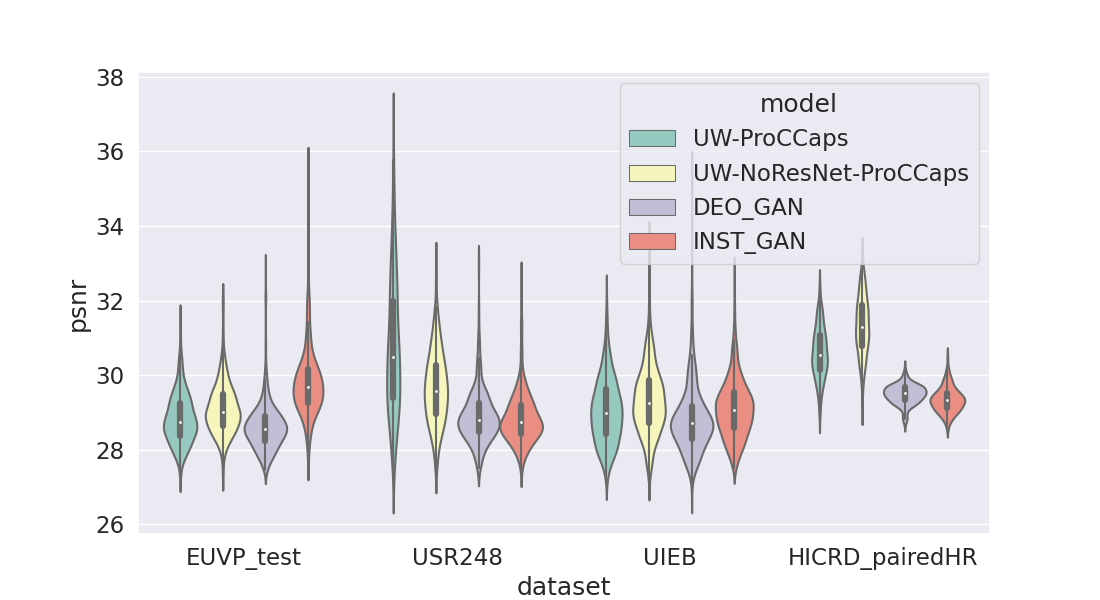}
\caption{Violin plot: compared results for the PSNR metric.}
\label{PSNR}
\end{figure}
We analyse the results obtained for the metrics presented in Sec.~\ref{sec4.3}. In Tabs.~\cref{tab1,tab2} the results with the proposed UW-ProCCaps and the Deodify~\cite{DeOldify} and InstanceAware~\cite{Su2020CVPR}. The results in Tab.~\ref{tab1} are results obtained with the models trained end-to-end with UFO120 dataset. We note that the UW-ProCCaps model outperforms the results obtained with the two models at the stat-of-the-art. This proves that the UW-ProCCaps model is a promising model that reaches good results already at the first stage of training. In Tab.~\ref{tab2}, the models are fine-tuned in GAN on UFO120. The results summarised in the tables are also visible for UW-ProCCaps and for Deodify and InstanceAware in Fig.~\cref{LPIPS,SSIM,PSNR} to facilitate the interpretation of the performances. For the PSNR and SSIM metrics, the higher result is better, while for the LPIPS metric, the lower result is better. As shown in the tables, the UW-ProCCaps outperforms the SOTA models with almost all the datasets. Finally, in Tab.~\ref{tab3}, we compare the UW-ProCCaps model with the enhancement models. The results obtained with our model outperform all the considered models. This proves that the final model obtained is robust and obtains high performances in the colourisation task and in the enhancement task.

\paragraph{Qualitative comparison}\label{sec5.2}
\begin{figure}
\begin{center}
\includegraphics[]{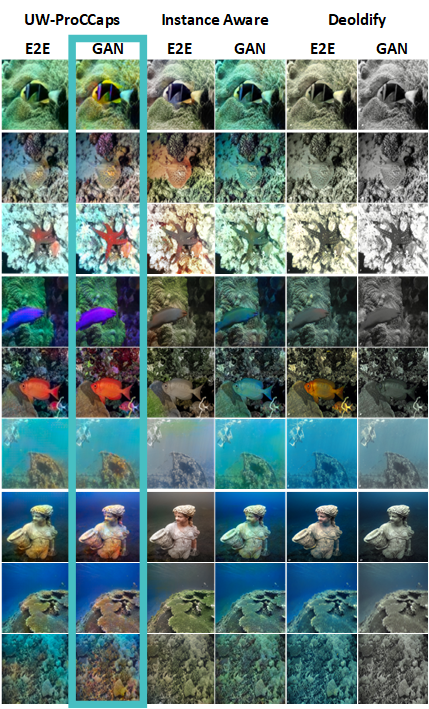}
\caption{Figure shows the colours obtained with the proposed UW-ProCCaps at the end-to-end and GAN phase and compares these results with colours reconstructed with models at the stat-of-the-art, Deodify~\cite{DeOldify} and InstanceAware~\cite{Su2020CVPR}.}
\label{figfinal}
\end{center}
\end{figure}
In Fig.~\ref{figfinal}, we summarise some samples of colourations obtained with the proposed model and the stat-of-the-art in both the end-to-end and GAN training phases. We think that presenting both the end-to-end and GAN results demonstrate that the model has a robust reconstruction of underwater colourisation while this behaviour is not obtained with the other models in the same conditions. We present in the UW-ProCCaps columns, the colour reconstructed with our model on validation datasets. The InstanceAware and the Deoldify columns are the colours reconstructed with models at the stat-of-the-art. The colours obtained with the UW-ProCCaps in GAN column are plausible and they present high quality on the detailed entities. The model is able to reconstruct different colourisations of the same entity as shown for the yellow-black fish, the purple fish, and the red fish in Fig.~\ref{figfinal}, and for the sea colours. The third row from the top, the starfish, present a complex case for colourisation because the model has to colourise the starfish against the seabed that is in the background. The UW-ProCCaps deals with the starfish providing a vibrant red colour while the stat-of-the-art is not providing a colourisation.

\section{Conclusion}\label{sec6}
In this paper, we take into consideration the colourisation of greyscale images for underwater domain. Taking into consideration the greyscale images we level up the cameras of colour resolution and colour distortion common underwater. Moreover the compression of each image to only the luminescence channel, reduce the memory space required for the image collection campaigns. The proposed UW-ProCCaps brings together different architectural strategies such as capsules, parallel classifier, and encoder-decoder structure; different training methodologies such as ProGL, GAN. We presented the ablation study of each on the main choices that we made for this model proving that each piece is doing is part to obtain the best balance between quantitative and qualitative performances. The results obtained are qualitatively superior to the stat-of-the-art showing bright colourisation and images with high quality. The quantitative results on metrics outperform the ones at the stat-of-the-art in the same training conditions. 
\bibliographystyle{IEEEtran}
\bibliography{manuscript}

\end{document}